\documentclass{article}
\pdfoutput=1
\usepackage{arxiv}
\usepackage{lineno,hyperref}
\usepackage{booktabs} 

\usepackage{amssymb}
\usepackage{amsthm}
\usepackage{amsmath}

\usepackage{hyperref}

\usepackage{algorithm}
\usepackage{algpseudocode}

\usepackage{tikz}
\usetikzlibrary{plotmarks}
\usepackage{pgfplots}
\usepackage{svg}
\usetikzlibrary{positioning,shapes,arrows,shadows,patterns}

\usepackage{multirow}
\usepackage{colortbl}

\usepackage{amsmath,amsfonts}
\usepackage{array}
\usepackage[caption=false,font=normalsize,labelfont=sf,textfont=sf]{subfig}
\usepackage{textcomp}
\usepackage{stfloats}
\usepackage{url}
\usepackage{verbatim}
\usepackage{graphicx}

\title{Car Sensors Health Monitoring by Verification Based on Autoencoder and Random Forest Regression}

\author{Sahar Torkhesari \\
  Department of Mathematics and Computer Science \\
  Amirkabir University of Technology\\ (Tehran Polytechnic)\\
  Iran\\
  \texttt{s.hesari@aut.ac.ir}\\
   \And
Behnam Yousefimehr \\
  Department of Mathematics and Computer Science \\
  Amirkabir University of Technology\\ (Tehran Polytechnic)\\
  Iran\\
  \texttt{behnam.y2010@aut.ac.ir}\\
   \And
  Mehdi Ghatee\footnote{The corresponding author} \\
  Department of Mathematics and Computer Science \\
  Amirkabir University of Technology\\ (Tehran Polytechnic)\\
  Iran\\
  \texttt{ghatee@aut.ac.ir}\\
}

\begin{document}
\maketitle
\begin{abstract}
Driver assistance systems provide a wide range of crucial services, including closely monitoring the condition of vehicles. This paper showcases a groundbreaking sensor health monitoring system designed for the automotive industry. The ingenious system leverages cutting-edge techniques to process data collected from various vehicle sensors. It compares their outputs within the Electronic Control Unit (ECU) to evaluate the health of each sensor. To unravel the intricate correlations between sensor data, an extensive exploration of machine learning and deep learning methodologies was conducted. Through meticulous analysis, the most correlated sensor data were identified. These valuable insights were then utilized to provide accurate estimations of sensor values. Among the diverse learning methods examined, the combination of autoencoders for detecting sensor failures and random forest regression for estimating sensor values proved to yield the most impressive outcomes. A statistical model using the normal distribution has been developed to identify possible sensor failures proactively. By comparing the actual values of the sensors with their estimated values based on correlated sensors, faulty sensors can be detected early. When a defective sensor is detected, both the driver and the maintenance department are promptly alerted. Additionally, the system replaces the value of the faulty sensor with the estimated value obtained through analysis. This proactive approach was evaluated using data from twenty essential sensors in the Saipa's Quick vehicle's ECU, resulting in an impressive accuracy rate of 99\%.

\end{abstract}

\keywords{Driver assistant system  \and Machine learning  \and Sensor fault diagnosis  \and Smart car  \and Autoencoder  \and Random forest regression}

\section{Introduction}
\label{Introduction}
In the realm of contemporary transportation, the prevalence of driver assistance systems in smart vehicles has witnessed significant growth. These systems are integral to maintaining the overall health and performance of vehicles \cite{dereszewski2019diagnostics}. Concurrently, the advancement of electronic technology has enabled not only the proliferation of diverse sensor types but also the facilitation of data processing from the vast quantities generated by these sensors. Nevertheless, as automotive systems increase in complexity, the challenge of ensuring the proper functioning of these sensors simultaneously intensifies. Even minor deviations or failures in sensor readings can lead to considerable repercussions \cite{li2023light}.

Current engineering practices focus on the development of precise sensors to monitor critical hardware components. Furthermore, innovative methodologies are routinely proposed to identify sensor-related issues and to leverage sensors in the online assessment of driver behavior, thus contributing to the maintenance of system performance. Internal combustion engines exemplify complex systems characterized by a multitude of sensors and control mechanisms \cite{mehranbod2005method}.

In such engines, the electronic control unit plays a pivotal role in regulating fuel injection and ignition timing to ensure an optimal fuel-air mixture for combustion. By continuously monitoring system variables, including engine airflow, throttle position, and engine speed, the electronic control unit is able to compute the appropriate fuel flow and ignition timing, thereby maximizing torque output. However, sensors may fail for various reasons, resulting in incorrect readings that compromise the effectiveness of the control unit. The failure of one or more sensors can precipitate inefficient operation, instability, or, in severe cases, engine failure \cite{suwatthikul2010fault}.

This paper addresses the identification of sensor errors through machine learning techniques such as Autoencoder \cite{hinton2006reducing} and estimates the values reported by sensors. In the event of a failure, the system is designed to detect such anomalies and substitute the erroneous readings with close-to-actual estimates derived from other operational sensors. The innovative aspect of this approach lies in the deployment of autoencoder networks for the estimation of vehicle sensor values, employing a threshold based on Gaussian distribution to assess the health status of various vehicle sensors—a concept introduced for the first time in this context. This pioneering methodology facilitates the detection of sensor failures without necessitating additional hardware and holds the potential to reduce the number of sensors required in future automotive engineering designs. The subsequent sections will review related works, present the proposed methodology, and evaluate the performance of the suggested system.

\begin{table}
\footnotesize
\centering
\label{tab:PW}
\caption{Previous works}
\begin{tabular}{|c|c|p{5cm}|p{5cm}|}
\hline
\textbf{Ref} & \textbf{Year} & \textbf{Error detection and identification method} & \textbf{Error} \\ \hline
\cite{amin2022unified} & 2022 & Fault detection and identification system with fuel-air ratio control. Triple modular hardware redundancy for sensors and dual redundancy for actuators. & Defects in exhaust gas re-circulation sensor, speed sensor, throttle position sensor, and air mass flow sensor \\ \hline
\cite{amin2023robust} & 2023 & Fault detection and identification system with high gain passive control, dual hardware redundancy for multiple failures & Defects in exhaust gas re-circulation sensor, speed sensor, throttle position sensor, and air mass flow sensor \\ \hline
\cite{amin2019advanced} & 2019 & Fault detection and identification system with high gain passive control, dual hardware redundancy for multiple failures & Defects in exhaust gas re-circulation sensor, speed sensor, throttle position sensor, and air mass flow sensor \\ \hline
\cite{shahbaz2021design} & 2021 & Active fault-tolerant control using a neural network-based nonlinear observer & Defects in exhaust gas re-circulation sensor, speed sensor, throttle position sensor, and air mass flow sensor \\ \hline
\cite{guzman2020fault} & 2020 & Fault detection and identification system based on neural network & Error in the sensor of the absolute pressure of the manifold and the mass flow of the air and the throttle position \\ \hline
\cite{mofleh2020fault} & 2020 & Fault detection and identification system based on audio signals and neural network & Ignition failure of cylinders 1 and 2, actuator failure \\ \hline
\cite{ghazaly2022prediction} & 2022 & Fault detection and identification system based on an unsupervised vibration algorithm that uses a neural network with a competitive learning algorithm & Engine ignition error \\ \hline
\cite{wang2020engine} & 2020 & Fault detection and identification system based on engine noise, sound intensity analysis techniques, incomplete wavelet packet analysis, and neural network & Malfunctions in cylinders, hall sensor, throttle orientation potentiometer, and exhaust gas recirculation sensor \\ \hline
\cite{cervantes2023multiple} & 2023 & Fault detection and identification system based on neural network & Error in sensors of throttle position and manifold air pressure and air mass flow \\ \hline
\end{tabular}
\normalsize
\end{table}

\section{Related work}
\label{RW}
The advancement of driver-assistance systems is pivotal for enhancing automotive quality and health measurement. The challenge of predicting sensor failures, alongside the associated repair or replacement processes, constitutes a complex issue, with preventive troubleshooting emerging as a significant concern in the domain of vehicle health monitoring systems. This innovative approach enables the identification of faulty sensors and their timely replacement prior to the occurrence of potential accidents. 

In the context of diagnosing sensor failures in automobiles, a variety of methodologies have been formulated, leveraging expert knowledge, detection mechanisms, and artificial intelligence. For instance, in \cite{murtaza2018super}, an adaptive control-based error detection method was developed specifically for turbochargers, diesel engines, and exhaust gas recirculation systems. Additionally, \cite{carbot2019ekf} introduced a fault detection system utilizing the Kalman filter focused on pressure and temperature sensors within the inlet air manifold of internal combustion engines. Moreover, \cite{amin2019robust} demonstrated an active fault-tolerant control strategy for air-fuel ratio management in internal combustion engines, utilizing a linear regression-based observer for detection, reconfiguration, and feedback control to uphold the air-fuel ratio. In a similar vein, \cite{li2021friction} presented a fault tolerance control strategy for electronic gas valves through the application of adaptive neural network estimators. Furthermore, Reference \cite{shahbaz2021design} proposed a fault-tolerant active control system grounded in artificial neural networks for managing the fuel-air ratio in spark-ignition engines, showcasing system stability even amid operational failures. The collective efforts of the authors in \cite{amin2022unified, amin2023robust, amin2019advanced} have also highlighted the detection of faults in sensors and actuators through hardware redundancy mechanisms, thereby activating fault-tolerant control in motors equipped with multiple controllers. In addition, references \cite{guzman2020fault, mofleh2020fault, yu2014dynamic, sangha2005fault, ghazaly2022prediction, wang2020engine, cervantes2023multiple} have employed various machine learning techniques to identify errors across diverse sensors and actuators.

An analysis of the prior works summarized in Table \ref{tab:PW} indicates a scarcity of research focused on diagnosing multiple faults within internal combustion engines. Previous studies examining multiple failures have predominantly advocated for the implementation of hardware redundancy in sensors; however, this approach entails significant costs. Additionally, the data utilized in these investigations were primarily gathered in controlled laboratory settings under constant environmental conditions, and were often limited to a small array of sensors. This limitation can result in models that may perform inadequately under real-world conditions, as they frequently overlook the intercommunication among multiple sensors.

In the current research, we utilize real-world data obtained while driving under varying environmental conditions. Furthermore, we employ an autoencoder artificial neural network \cite{hinton2006reducing} to analyze internal combustion engine sensors and elucidate the relationships among different features for error detection purposes. Unlike many preceding studies which considered a limited number of sensors, categorizing some as features and others as objective functions, the proposed system overcomes this limitation by collectively considering the interrelationships of all features.

\section{Car Sensors Health Monitoring System}
\label{Methodology}
\begin{figure}
\centering
\includegraphics[scale=0.371]{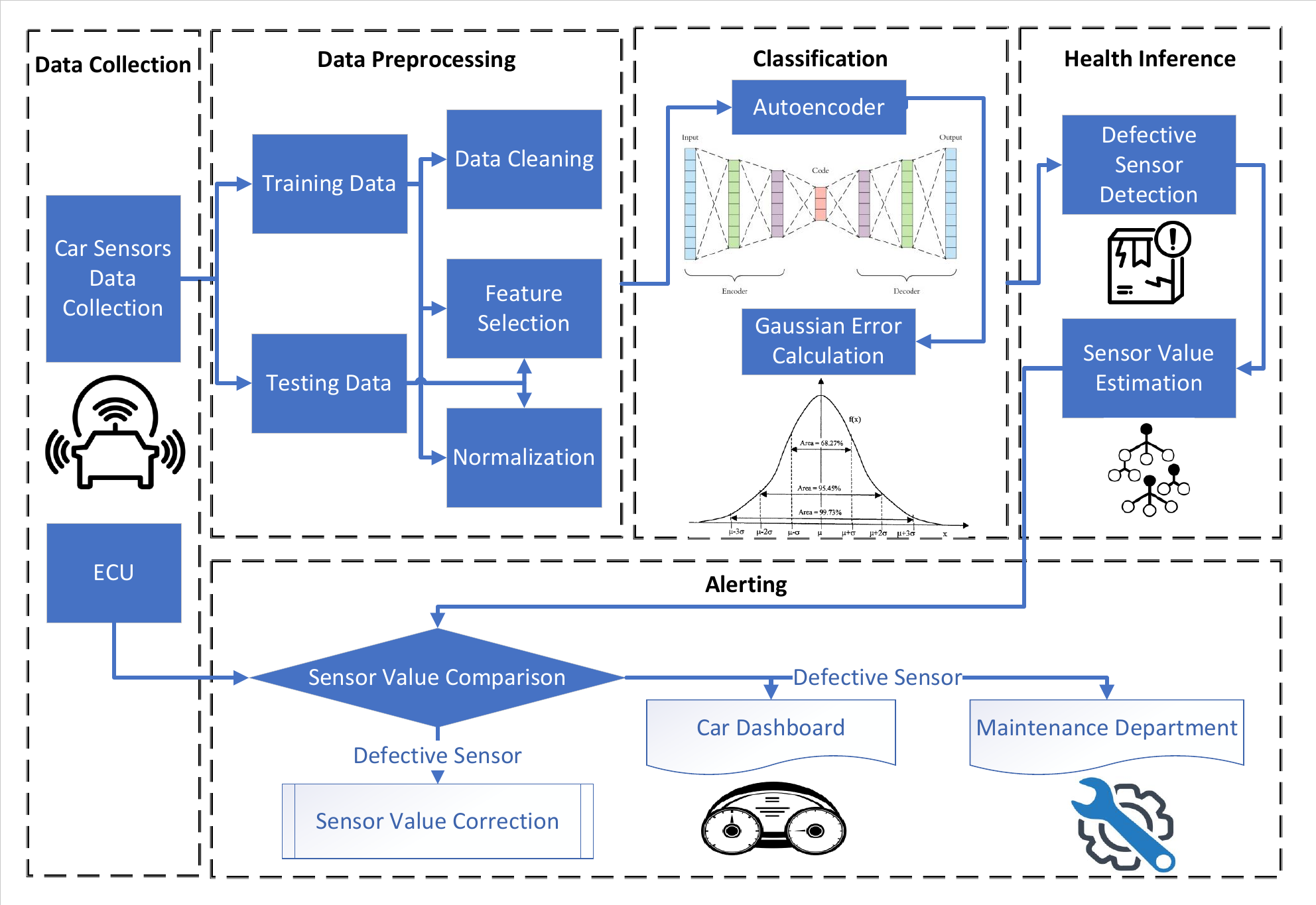}
\caption{Car Sensors Health Monitoring System}
\label{fig:Arch}
\end{figure}

In this section, we introduce a novel system designed to assess the health of vehicle sensors. The architecture of this system is illustrated in Figure \ref{fig:Arch}. This system is tasked with evaluating the condition of vehicle sensors, identifying defective sensors, and substituting erroneous values with estimated predictions derived from other correlated sensors.

Initially, this system collects data and then performs necessary pre-processing steps. Subsequently, the autoencoder neural network \cite{hinton2006reducing} is employed to predict sensor values. It then computes the difference between the estimated model value and the sensor output, comparing it against predetermined values for each sensor. The process categorizes sensor status into different classes, including healthy, almost healthy, normal, almost defective, and defective. By using the random forest regression model \cite{breiman2001random}, the system can detect and substitute defective sensor values for better or worse outcomes. In what follows, a more detailed explanation of the operation and components of each step is given.
\subsection{Data Collection}
SaipaYadak company has initiated a collaborative project with Irancell operator and Apadana Dolphin company to develop connected cars. Dolphin Apadana company has designed boards that connect to the electronic control unit via the debugger socket, retrieving data from a range of sensors at intervals ranging from one to ten minutes. This data is transmitted through the SIM card installed on Saipa company's dedicated servers. The system facilitates the transmission, storage, and display of vehicle information including geographic location, speed, and error logs from the car's electronic control unit. This functionality is accessible through mobile software developed by Apadana Dolphin Company and installed on the driver's phone. Previously, access to electronic control unit errors was limited to diagnostic devices.

This research primarily focuses on utilizing the internal combustion engine of Quick cars. By leveraging the aforementioned system, data from various sensors within the car's electronic control unit is collected, recorded, and utilized.
\subsection{Preprocessing}
In this section, we begin by partitioning the dataset, allocating 33\% for evaluation and 67\% for training. To establish an error threshold, 25\% of the training subset is further separated for validation.\\
Following data separation, we proceed with data cleansing and normalization. Data cleansing involves the identification, removal, or correction of defective or incorrect records within the dataset. This process entails identifying incomplete, incorrect, or irrelevant data segments and subsequently replacing, correcting, or deleting them.\\
In the proposed approach, the dataset is initially assessed for the presence of noise, which is then eliminated if detected. Next, features about the sensors are selected from the entire dataset, and finally, the data is normalized. Data normalization standardizes the range of features within the dataset, a crucial step for machine learning algorithms, as they exclusively operate on numerical inputs. Variations in data value ranges can potentially skew the learning process and yield erroneous assumptions. In this study, mean-max normalization is employed to mitigate this issue.
\subsection{Classification}
\begin{figure}
\centering
\includegraphics[]{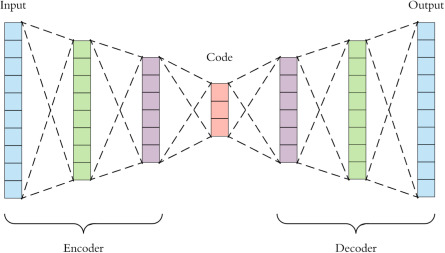}
\caption{Autoencoder\cite{ABIRAMI2020339}}
\label{fig:Autoencoder}
\end{figure}

To detect defective sensors, we employ an autoencoder neural network \cite{hinton2006reducing}, whose architecture is illustrated in Figure \ref{fig:Autoencoder}.
An autoencoder is a type of artificial neural network architecture utilized for learning compact representations. Typically, this model comprises two primary components: an encoder and a decoder \cite{hinton2006reducing,breiman2001random,daneshfar2024elastic}.

\begin{itemize}
\item \textbf{Encoder:} Responsible for converting input into a hidden space, this component compresses information from higher dimensions to lower dimensions.
\item \textbf{Decoder:} Receiving information from the hidden space, the decoder endeavors to reconstruct the information into the primary dimensions of the input.
\end{itemize}
In estimating the error, we treat sensor data as both the input and output of the model. We expect the model to learn the interrelationships among sensors through the autoencoder's middle layer, thereby constructing the output layer based on this learned representation. By comparing the actual sensor values with those predicted by the model, we ascertain the sensor's health status.

To this end, twenty of the most interconnected sensors are selected as input for the autoencoder. Utilizing an intermediate layer comprising twelve neural nodes, the data is mapped to a lower dimension, facilitating intermediate layer production. After generating data in the output layer, the goal is to minimize the disparity between this data and the initial input. By measuring the distance between these variables, sensor errors can be identified. The detection coefficient for each feature is then calculated in Table \ref{tab:RAE} based on the accuracy obtained for each sensor.

\begin{table}[]
\footnotesize
\centering
\caption{R-squared coefficient obtained for different features using Autoencoder model}
\label{tab:RAE}
\begin{tabular}{|l|l|}

\hline
Feature                                          & R\textasciicircum{}2 (R-squared correlation) \\ \hline
The temperature of the air entering the manifold & 0.99985                                      \\ \hline
Pressure inside the manifold                     & 0.999953                                     \\ \hline
Stepper rotation rate                            & 0.85195                                      \\ \hline
Engine speed                                     & 0.99972                                      \\ \hline
Throttle position sensor voltage                 & 0.883058                                     \\ \hline
fuel injection time                              & 0.989544                                     \\ \hline
Throttle position                                & 0.904884                                     \\ \hline
Engine water temperature                         & 999915                                       \\ \hline
Coil charging time                               & 0.358985                                     \\ \hline
Battery voltage                                  & 0.999985                                     \\ \hline
Vehicle condition                                & 0.999925                                     \\ \hline
upstream oxygen voltage                          & 0.999976                                     \\ \hline
downstream oxygen voltage                        & 0.9999                                       \\ \hline
Speed                                            & 0.99975                                      \\ \hline
Percentage of load on the motor                  & 0.165281                                     \\ \hline
Canister percentage                              & 0.99989                                      \\ \hline
Fan status                                       & 0.99989                                      \\ \hline
Advance angle                                    & 0.99993                                      \\ \hline
Move                                             & 1                                            \\ \hline
Strike                                           & 0.99986                                      \\ \hline
\end{tabular}
\normalsize
\end{table}

\begin{figure}[h]
    \centering
    \includegraphics[width=0.5\textwidth]{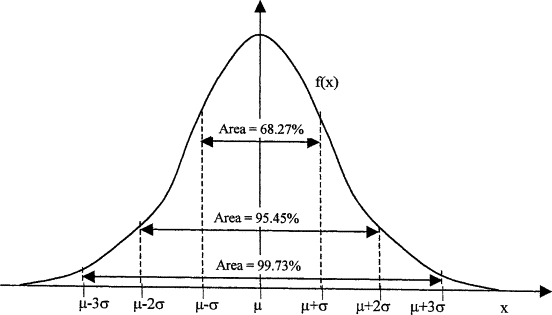}
    \caption{Illustration of standard deviation's effect on data distribution \cite{Gulati2022}.}
    \label{fig:std_dev}
\end{figure}

To assess the magnitude of error, we undertake the following steps:

\begin{enumerate}
\item After receiving data from sensors (actual data) and processing it through the autoencoder model, we derive the estimated value from the model. This is accomplished by calculating the absolute difference between the actual value and the estimated value, which reflects the discrepancy between these two variables.
   
\item Assuming that the error data follows a normal distribution, we compute the standard deviation based on the validation data and use it as a threshold for error classification.
\end{enumerate}
The standard deviation serves as a measure of dispersion, indicating how far data points deviate from the mean. As illustrated in Figure \ref{fig:std_dev}, an increase in the standard deviation corresponds to a smaller percentage of data within that region of the distribution, thereby heightening the likelihood of errors.

As a result, after calculating the standard deviation and data distance from the mean, the health of the sensors according to Table \ref{tab:health_index} are classified into healthy, almost healthy, normal, almost defective and defective classes.

\begin{table}[htbp]
\footnotesize
    \centering
    \caption{Health Index Categories Based on Value Difference ($E$)}
    \label{tab:health_index}
    \begin{tabular}{ll}
        \toprule
        \textbf{Value Difference ($E$)} & \textbf{Health Index} \\
        \midrule
        $µ-\sigma < E < µ+\sigma$ & Healthy \\
        $µ\pm\sigma \leq E \leq µ\pm2\sigma$ & Almost healthy\\
        $µ\pm2\sigma \leq E \leq µ\pm3\sigma$ & Normal \\
        $µ\pm3\sigma \leq E \leq µ\pm4\sigma$ & Almost defective \\
        $E > µ\pm4\sigma$ & Defective\\
        \bottomrule
    \end{tabular}
\end{table}
\subsection{Health Inference}
When a sensor failure is identified, simply relying on the value estimated through the autoencoder network may not suffice. The inaccuracies inherent in the autoencoder model necessitate a more robust approach. In such cases, a combination of techniques involving random forest regression proves effective [22]. By integrating insights from both the autoencoder network and the random forest regression model, along with leveraging data from other sensors as features and considering inter-sensor correlations, a more accurate estimation of the defective sensor's value can be achieved.

It is essential to note that the random forest model undergoes separate training, occurring after the pre-processing stage. This ensures that the model is finely tuned to the specific task at hand. Table \ref{tab:RFR} provides an overview of the additional sensor features required to estimate the value of each sensor accurately.\\
Through this amalgamation of methodologies, a more reliable estimation of the defective sensor's value can be obtained, thereby facilitating timely and effective replacements, ultimately contributing to enhanced system performance and reliability.
\begin{table}
\centering
\caption{Random forest regression evaluation table for each sensor}
\label{tab:RFR}
\begin{tabular}{|p{5cm}|p{2cm}|p{2cm}|p{2cm}|}
\hline
Property                                         & Number of appropriate features & Mean absolute error & R-squared correlation \\ \hline
The temperature of the air entering the manifold & 19                             & 0.020632            & 0.978438              \\ \hline
Pressure inside the manifold                     & 6                              & 0.002668            & 0.998619              \\ \hline
Stepper rotation rate                            & 3                              & 0.013861            & 0.98679               \\ \hline
engine speed                                     & 5                              & 0.010328            & 0.99039               \\ \hline
Throttle position voltage                        & 4                              & 0.00127             & 0.999277              \\ \hline
fuel injection time                              & 7                              & 0.00588             & 0.99253               \\ \hline
Throttle position                                & 5                              & 0.00087             & 0.999141              \\ \hline
Engine water temperature                         & 18                             & 0.02064             & 0.989771              \\ \hline
Coil charging time                               & 2                              & 3.5244*10-5         & 0.999736              \\ \hline
Battery voltage                                  & 6                              & 0.00612             & 0.99844               \\ \hline
Vehicle condition                                & 17                             & 0                   & 1                     \\ \hline
upstream oxygen voltage                          & 17                             & 0.173308            & 0.56299               \\ \hline
downstream oxygen voltage                        & 17                             & 0.12036             & 0.544329              \\ \hline
Speed                                            & 15                             & 0.050812            & 0.8361                \\ \hline
Percentage of load on the motor                  & 9                              & 0.00683             & 0.994099              \\ \hline
Canister percentage                              & 2                              & 0.073999            & 0.831502              \\ \hline
Fan status                                       & 5                              & 0.021867            & 0.92666               \\ \hline
Advance angle                                    & 3                              & 0.024456            & 0.965761              \\ \hline
Move                                             & 1                              & 0                   & 1                     \\ \hline
strike                                           & 19                             & 0.10012             & 0.68443               \\ \hline
\end{tabular}
\end{table}
\section{Evaluation}
The proposed system was evaluated using the coefficient of determination, a statistical measure commonly employed in regression analysis to assess the alignment of the regression model with actual data. This coefficient ranges between zero and one, with higher values indicating a stronger fit of the model to the data.
The formula for the coefficient of determination is:
\begin{equation}
R^2 = 1 - \frac{\sum_{i=1}^{n}(y_i - \hat{y}_i)^2}{\sum_{i=1}^{n}(y_i - \bar{y})^2}
\end{equation}

where:
\begin{itemize}
    \item \( R^2 \) is the coefficient of determination.
    \item \( y_i \) represents the actual observed values in the data.
    \item \( \hat{y}_i \) represents the values predicted by the regression model.
    \item \( \bar{y} \) is the average of the observed \( y \) values in the data.
\end{itemize}

Upon evaluating the Quick data, it was found that the coefficient of determination for all 20 sensors in the dataset exceeded 99\%. Table \ref{tab:Compresion} provides a comparison between the proposed system and recent related works. As can be seen, the proposed system has been able to achieve great accuracy on non-laboratory data and a large number of sensors with less computational complexity.

\begin{table}
\centering
\caption{Comparison of the proposed system with recent works}
\label{tab:Compresion}
\begin{tabular}{|l|l|p{2cm}|p{2cm}|p{2cm}|}
\hline
Ref & Year & Number of sensors & Number of models & R-squared correlation \\ \hline
\cite{amin2019robust} & 2019 & 4 & 4 & 0.8 \\ \hline
\cite{yu2014dynamic} & 2020 & 2 & 3 & 0.93 \\ \hline
\cite{amin2022unified} & 2023 & 3 & 5 & 0.99 \\ \hline
Our method & 2023 & 20 & 2 & 0.99 \\ \hline
\end{tabular}
\end{table}

\section{Conclusion and Future Work}
This article introduces a novel system designed to predict and assess failures in car sensors. The system integrates an autoencoder neural network with random forest regression, providing a robust architecture capable of processing sensor data inputs to determine the health index for each sensor. By employing a predefined error threshold and leveraging the autoencoder model's outcomes, the system identifies potential sensor failures, substituting estimates for faulty sensors through a random forest regression model. Notably, the system's development and evaluation utilized real-world vehicle data collected by SaipaYadak.\\
In terms of future research directions, it is suggested to conduct a more comprehensive exploration of the proposed failure threshold for health indicators. This entails refining the accuracy of the failure threshold across various sensors, thereby enhancing the system's predictive capabilities and ensuring its effectiveness in real-world scenarios. Such endeavors would contribute significantly to advancing sensor failure prediction and mitigation strategies within automotive environments.

\section{Acknowledgment}
The authors extend their sincere gratitude to SaipaYadak Company and Dolphin Apadana Company for generously providing the data essential for this research. Their invaluable contribution and commitment to enhancing automotive technologies have been instrumental in the advancement of this study.




\bibliography{sn-bibliography}



\end{document}